\documentclass[letterpaper, 10 pt, conference]{ieeeconf}
\IEEEoverridecommandlockouts                              
\usepackage{cite}
\usepackage{amsmath,amssymb,amsfonts}
\usepackage{algorithmic}
\usepackage{textcomp}
\usepackage{xcolor}
\usepackage[pdftex]{graphicx}
\graphicspath{{images/}}
\DeclareGraphicsExtensions{.pdf,.jpeg,.jpg,.png}
\usepackage{mathptmx} 
\usepackage{times} 
\usepackage{amsmath} 
\usepackage{amssymb}  
\usepackage{hyperref}

\def\BibTeX{{\rm B\kern-.05em{\sc i\kern-.025em b}\kern-.08em
    T\kern-.1667em\lower.7ex\hbox{E}\kern-.125emX}}

\title{\LARGE \bf PatchMatch-Stereo-Panorama, a fast dense reconstruction from 360° video images}

\author{Hartmut Surmann$^{1}$, Marc Thurow$^{1}$, Dominik Slomma$^{2}$
\thanks{$^{1}$University of Applied Science, Gelsenkirchen, $^{2}$German Rescue Robotic Centre, Dortmund.}
\thanks{corresponding author: {\tt\small hartmut.surmann@w-hs.de}}
}

\begin{document}

\maketitle

\begin{abstract}
This work proposes a new method for real-time dense 3d reconstruction for common 360° action cams, which can be mounted on small scouting UAVs during USAR missions. The proposed method extends a feature based Visual monocular SLAM (OpenVSLAM, based on the popular ORB-SLAM) for robust long-term localization on equirectangular video input by adding an additional densification thread that computes dense correspondences for any given keyframe with respect to a local keyframe-neighboorhood using a PatchMatch-Stereo-approach. While PatchMatch-Stereo-types of algorithms are considered state of the art for large scale Mutli-View-Stereo they had not been adapted so far for real-time dense 3d reconstruction tasks. This work describes a new massively parallel variant of the PatchMatch-Stereo-algorithm that differs from current approaches in two ways:
First it supports the equirectangular camera model while other solutions are limited to the pinhole camera model. Second it is optimized for low latency while keeping a high level of completeness and accuracy. To achieve this it operates only on small sequences of keyframes, but employs techniques to compensate for the potential loss of accuracy due to the limited number of frames. Results demonstrate that dense 3d reconstruction is possible on a consumer grade laptop with a recent mobile GPU and that it is possible with improved accuracy and completeness over common offline-MVS solutions with comparable quality settings. 
\end{abstract}

{\bf keywords}: PatchMatch-Stereo, 360°-Panorama, visual monocular SLAM, UAV, Rescue Robotics

\section{Introduction}
\label{sec:introduction}
First responders have to make critical decisions to cope with disaster scenarios such as earthquakes, fires or floods \cite{Kruijff2014,Kruijff-amatrice}. 
Furthermore, the information about the situation is inaccurate and incomplete and decisions have to be taken under time pressure.
At the same time they regularly expose themselves to a high level of risk.
When searching for and rescuing trapped people from a building in danger of collapsing, the information must first be determined, where they are and what their state of health is. Only then can the actual rescue operation begin.
If rescuers enter the building to gather information and perform the rescue and a collapse occurs during the ongoing search, this also endangers the lives of the rescuers.

\begin{figure}
\centering
\includegraphics[width=0.49\textwidth]{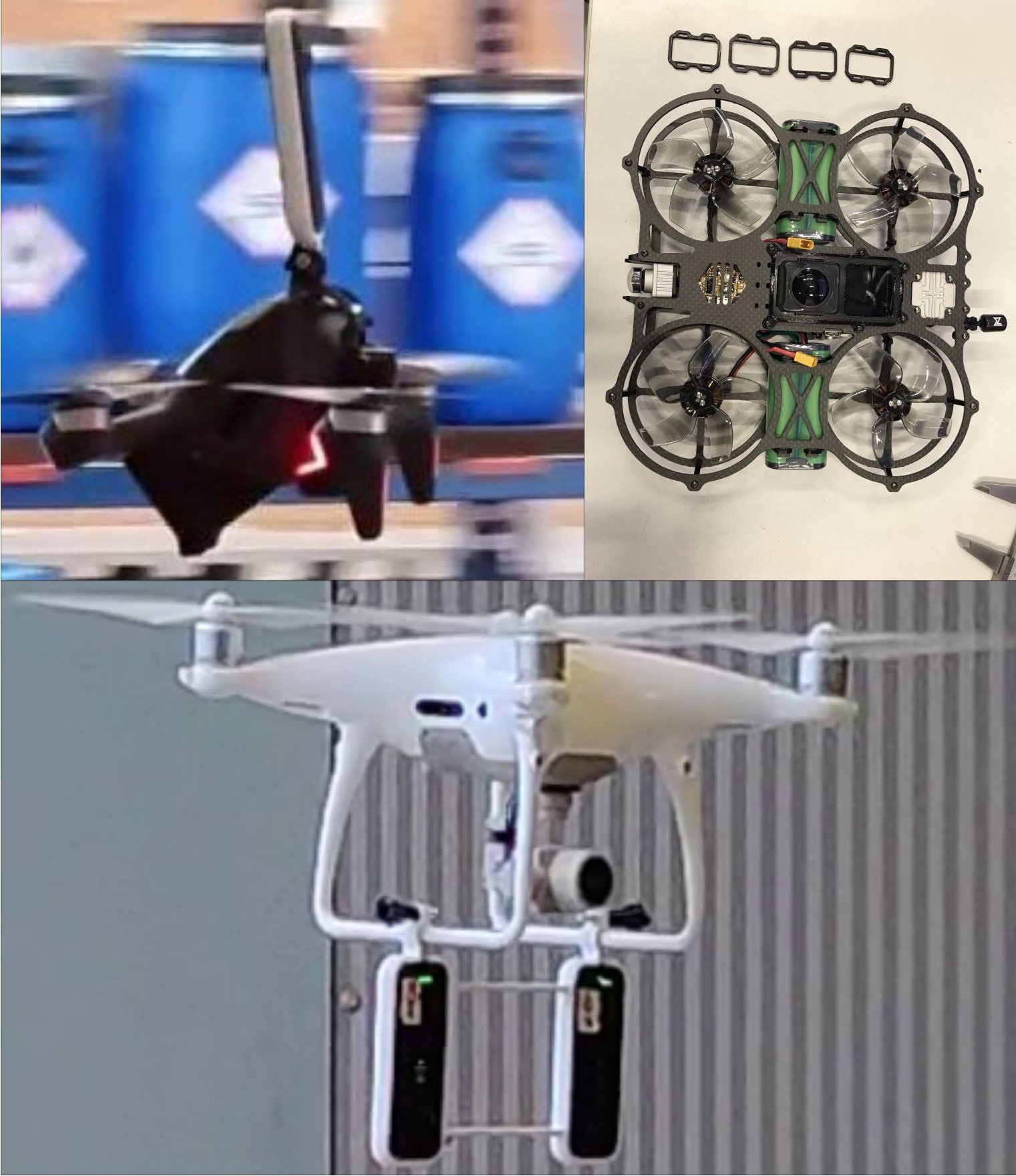}
\caption{Three different UAVs with 360° cameras. A DJI FPV with an Insta360 One X on top left. The invisible drone with an Insta360 one R in the top right and the Phantom P4 with two Insta360 One X at the bottom.}
\label{fig:uav360}
\end{figure}

Effective information gathering by robots promises to significantly minimize the risk to rescuers and increase the effectiveness of the rescue operation, such as during an operation following a flood disaster\cite{9738529}.
The range of capabilities of current robotic systems is still limited. However, individual robots usually have specialized forms of information gathering or specialized environmental manipulation capabilities.
But technical issues are not the only ones at play. 
Since robots are to be involved in critical decision-making processes, a holistic research and development process is needed that involves the rescue workers. 
For routine deployment, they must use the robots as an extension of their own range of capabilities.

Such a holistic approach is being within the framework of the joint research project establishment of the German Rescue Robotics Center (A-DRZ)\cite{9597869}.
It combines the competences of rescue services and fire departments with the robotics competences of different research groups and institutions. 
Realistic rescue scenarios are defined and the development of robotic systems is geared to these scenarios. At the same time, rescue forces get to know the application of robots.

Small ($\approx$35cm) and lightweight ($\approx$1kg) UAVs (Unmanned Aerial Vehicle) make an important contribution to the current rescue research\cite{9376551} and have been used in both TRADR \cite{Kruijff-amatrice} and A-DRZ \cite{9597677} for rapid indoor and outdoor reconnaissance. 
In both cases, commercial solutions were used, which are inexpensive and already have a wide range of capabilities.
These UAVs are teleoperated and their primary sensor is a high-resolution camera.
They can be used very flexibly. 
At high altitudes, they can autonomously fly over large areas and provide a rapid overview of the threat situation.
Nevertheless, due to their stable flight attitude, simple control and small size, they are suitable for the navigation in closed indoor spaces and take images.
3D models can be reconstructed from these images, which help the rescuers to assess the building conditions.

A major disadvantage of image acquisition with the onboard cameras of UAVs is the limited field of view ($\approx$90°).
In the recent past, inexpensive and lightweight 360° action cams have appeared that are small and light enough to be mounted on the UAV.
These can store high-resolution 360° videos ($\approx$5.7k) at 30 fps on an internal
SD card, or stream a low-resolution video over a wireless interface. With the
panoramas, a complete 360° view is possible, which pushes the fast inspection of indoor spaces forward. 


There is also a great interest to obtain dense 3d data from scouting UAVs or UGVs (Unmanned Ground Vehicle) for the purpose of distance measurement and path planning for other (potentially heavier and bigger) robots following up. However, 3d data is not provided by panoramic images alone, but can be reconstructed with photogrammetric methods.

One can identify two types of applications: Monocular Visual SLAM methods for online-localization on live video and Multi-View-Stereo (MVS) dense reconstruction on large scale unordered image data sets. While SLAM is mostly concerned with accurate visual odometry (VO) and long term consistent mapping under poor image quality in real time, only sparse or semi-dense pointclouds are reconstructed as a byproduct. On the other hand, MVS-densification methods are optimized for high level of accuracy and completeness. Image details and quality might vary, but should generally be as good as possible to improve the reconstruction results. Memory and computational complexity is also of great interest to improve scalability for ever larger image datasets with ever larger image sizes. 

So far there is no method for real time dense 3d reconstruction on a panoramic video that might be recorded or streamed by any common 360° action cam during active operation e.g. in indoor search and rescue missions (e.g. mounted on a scouting UAV). Image quality might not be ideal in such a case: Rapid movements might blur the image. This, as well, can happen due to vibrations from the rotors of an UAV. Image resolution might need to be reduced to process data in time. Furthermore, textureless environments or repeating structures are very common and harden the reconstruction tasks.

This work describes a method to obtain a dense 3d reconstruction from monocular panoramic video in real time by combining the two domains of VSLAM and MVS: It relies on OpenVSLAM\cite{Sumikura_2019} for accurate and robust long-term tracking and localization of an equirectangular camera and extends this by a novel variant of PatchMatch-Stereo\cite{6409456}\cite{Bleyer2011PatchMatchS} that is optimized for low latency densification on small sequences of equirectangular keyframes. Results demonstrate that real time reconstruction is possible on a consumer grade laptop to which the panoramic video can either be streamed or transferred immediately after mission return. Examples can be found at:

\url{https://github.com/RoblabWh/PatchMatch}.

\section{Related work}

\begin{figure}
\centering
\includegraphics[width=0.49\textwidth]{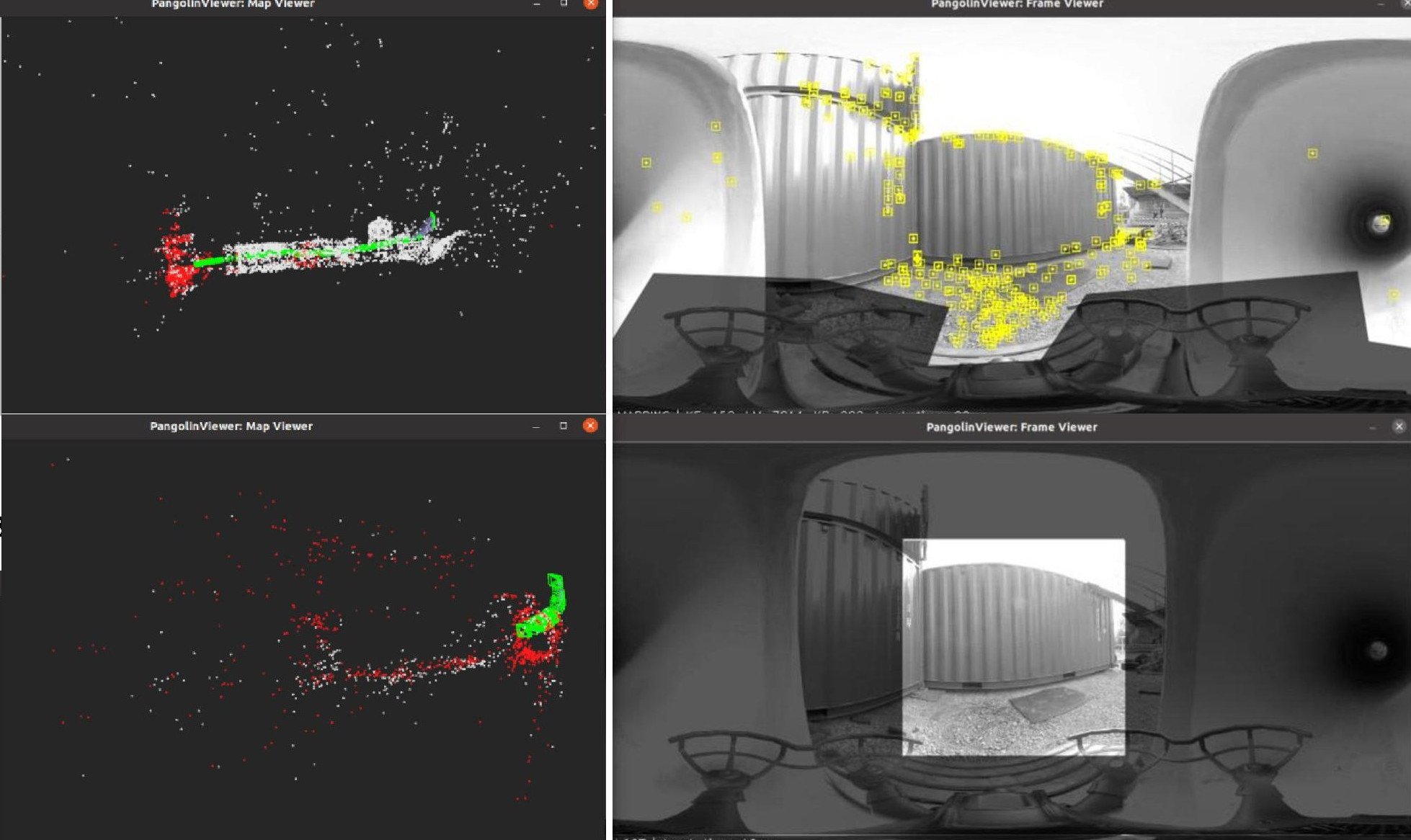}
\caption{Advantage of 360° camera images in difficult environments here a tube. The limit FOV of 90° leads to a tracking loss (bottom) whereas the 360° images enables stable tracking (top). }
\label{fig:advantageofequi}
\end{figure}

Monocular VSLAM is an active research field in mobile robotics. The methods include two distinct areas: Feature based and direct methods. Feature based methods attempt to match sparse, but robust features among multiple images and then derive position and orientation of these images and the positions of the features by minimizing a reprojection error with respect to a given camera model. 
ORB-SLAM\cite{Mur_Artal_2015} is one of the best known and successful representatives of this category with support for the pinhole camera model with limited field of view. Several iterations and derivatives have since emerged. ORB-SLAM2\cite{Mur_Artal_2017} is an extension of the original method that adds support for RGBD sensory and stereo camera systems. A publicly available implementation has been released under open source license. OpenVSLAM\cite{Sumikura_2019} is an independent derivative of the original ORB-SLAM that adds support for fisheye cameras and equirectangular 360° panoramas. It is a flexible open source framework that is easily expandable while having very much in common with the original ORB-SLAM. Its capability to process 360° panoramas though is an effective counter for tracking losses in feature less environments and under rapid camera movements as it often appears in real scenarios (see figure \ref{fig:advantageofequi}).The procedure in \cite{s21030705} is an extension of ORB-SLAM2 with a special focus on processing equirectangular 360° panoramas which ORB-SLAM3 did not have. Instead of still using ORB features as OpenVSLAM, it uses SPHORB (Spherical Oriented FAST and Rotated BRIEF) features, more adapt to equirectangular distortion. ORB-SLAM3\cite{Campos_2021} further extends ORB-SLAM2 by adding support for fisheye camera models, IMU sensor data fusion and multi-map capabilities. In conjunction with further enhanced relocation abilities, ORB-SLAM3 outperforms its predecessors and other comparable methods in terms of accuracy and long-term operational capabilities, including direct methods such as DSO\cite{DSO}.

Direct methods operate directly on pixel intensities and optimize the camera pose by minimizing a photometric error. These methods can be further divided into dense (all pixels), semi-dense (pixels with strong gradients) and sparse procedures (specially selected pixels with strong gradients). DPPTAM (Dense Piecewise Planar Tracking and Mapping from a Monocular Sequence)\cite{7354184} is a dense method that tries to reconstruct texture-rich regions with a semi-dense approach and low-texture regions by approximating surfaces. This procedure is real time capable on a CPU, but it is a pure VO method and thus suffers from accumulative errors. The best known semi-dense method are LSD-SLAM (Large-Scale Direct SLAM)\cite{10.1007/978-3-319-10605-2_54} and SVO (Semi-Direct-Visual Odometry)\cite{6906584}. The latter is a pure VO method and has inferior localization capability, but latency is very low even on a CPU. DSO (Direct Sparse Odometry)\cite{DSO} is a direct but sparse VO method. This method attempts to detect particularly robust pixel correspondencies among a number of consecutive keyframes. Similar to the indirect-feature-based SLAM method, a sparse point cloud is reconstructed in the process. DSO is as well capable of real time execution on a CPU. DSO and its numerous variants represent the current state of the art for direct VO methods. LDSO\cite{gao2018ldso} extends DSO with loop-closing capabilities and pose-graph optimization, better competing with recent feature based SLAM methods.
\begin{figure*}
\centering
\includegraphics[width=0.95\textwidth]{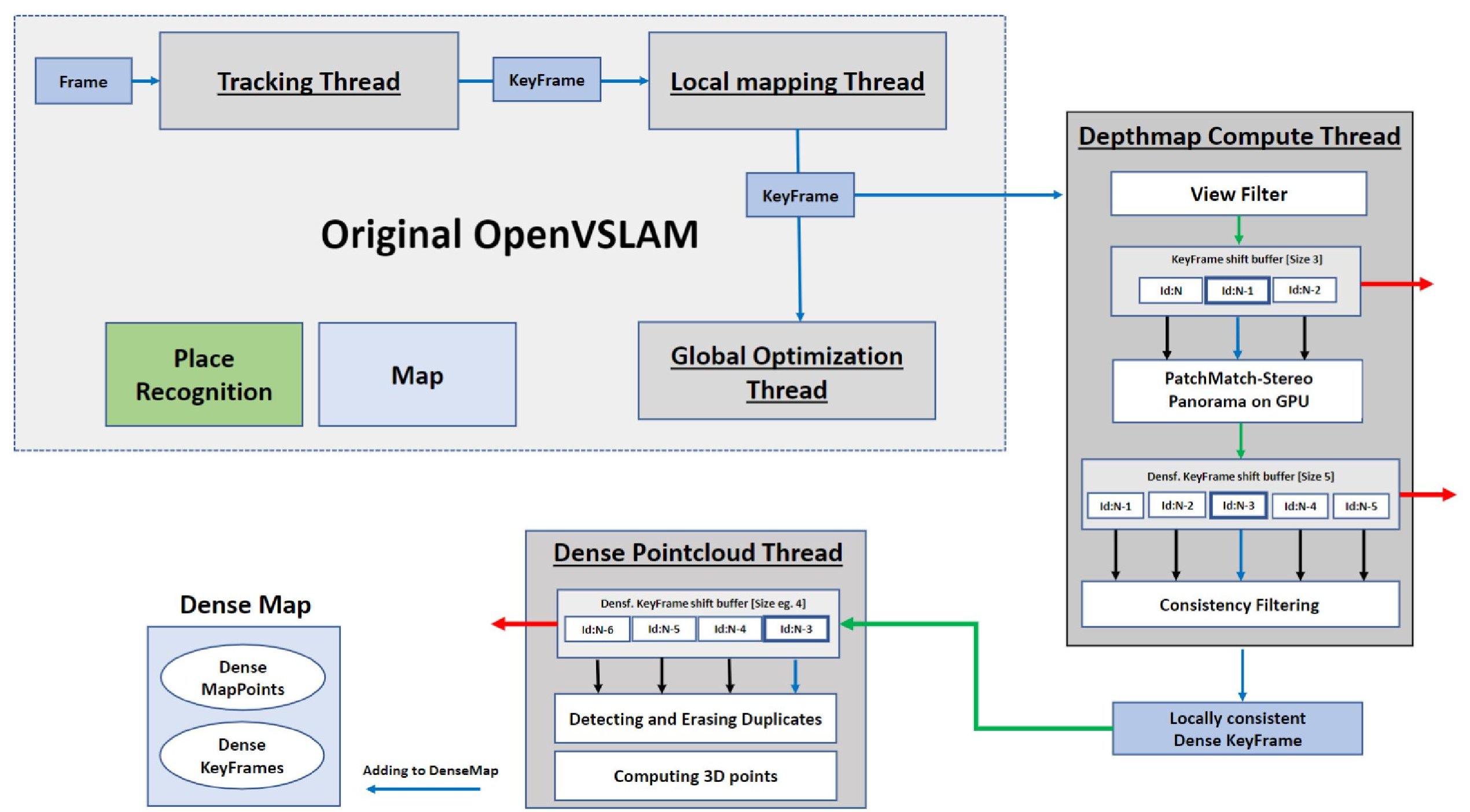}
\caption{Overview of the proposed approach. Top left shows the original OpenVSLAM algorithm whereas our extension is shown at the right and bottom part and starts with the blue line using the keyframes of OpenVSLAM }
\label{fig:overview}
\end{figure*}

\section{Proposed method}
The approach chosen here uses OpenVSLAM for the localization of the UAV (360° camera) and extends it with a 360° implementation of the PatchMatch-Stereo\cite{6409456}\cite{Bleyer2011PatchMatchS} algorithm to process three consecutive keyframes to a dense 360° depth image on the graphics card.
Figure \ref{fig:overview} shows the overview of the approach. Top left shows the original OpenVSLAM algorithm whereas our extension is shown at the right and bottom part and starts with the blue line using the keyframes of OpenVSLAM. 

It is a noteworthy point that OpenVSLAM can perform SLAM with an equirectangular camera and extends ORB2-SLAM. We start our explanation which a short review of the OpenVSLAM block.
With ORB-SLAM, each frame of the video is localized by a map containing reconstructed points, keyframes, and information about associated features.
Keyframes are a special selection of frames that allow a particularly robust feature assignment. Since effects with a negative influence on the image quality 
such as motion blur, overexposure or underexposure, noise or pronounced texture blur, the
the corresponding images are indirectly filtered in this way. Furthermore, with
that there is sufficient perspective variation for stereo reconstruction, since it is already ensured that these are suitable for localization. The number of keyframes is much smaller than the number of all frames of a video. Accordingly, the 
spatial and temporal distances between keyframes are higher, and the latency requirements for a dense a later densification component are correspondingly lower.
The selection of keyframes and their parameters are regularly updated during the mapping process to accommodate only those with high information content. The set of keyframes is kept sparse, the map as a whole contains comparatively few parameters, so that global optimization tasks can be performed. This concerns for example for the loop closing, which uses a trajectory with a loop shape to resolve inconsistencies between the start and end points of the loop by means of an error re-projection. If a frame is declared to be a keyframe, then an initial pose estimation and parameter estimation is already available. Furthermore, pose parameters, 3D points and further camera parameters for this keyframe are estimated based on a set of neighboring keyframes with common feature points by a bundle adjustment (BA) step. This process has a comparatively high latency, which is why ORB-SLAM and OpenVSLAM separate the initial position estimation in one tracking thread and the local BA for the selected key frame neighbors in a local mapping thread. The local mapping thread effectively has a stream of well localized 360° panoramas, which are presumably well suited for dense 3D reconstruction. 

The extended cuda implementation of the PatchMatch-Stereo algorithm for 360° images takes every keyframe of OpenVSLAM at separated depth map compute thread. The keyframes came out of OpenVSLAM with a much lower frame rate than the original 30 fps of a video. First, the keyframes are filtered and queued. If 3 frames are available a complete depth image is calculated and the resulted depth image is also queued. If five depth maps are available the current depth image is filtered with a consistency filter and a locally consistent dense frame is taken to a further dense point cloud thread. Due to the filtering outliers are removed and the resulting depth map is not complete any more. The dense point cloud thread buffers 4 locally consistent dense keyframes and erases duplicates. The resulting 3D points are added to the dense map.

So, lets take a closer look to the PatchMatch-Stereo and the filtering. Previous PatchMatch solutions only support the pinhole camera model. Current 3D reconstruction software like OpenDroneMap (ODM) \cite{odm2020} is able to reconstruct equirectangular panoramas.
In this case, the densification is done with a PatchMatch-Stereo algorithm, but with pinhole camera images. Therefore, panoramas are transformed into pinhole images by an equirectangular to cubemap transformation which contains six faces (cube faces). These 6 single images are treated as independent shots by the current PatchMatch-Stereo implementations. In the process of neighborhood determination to the images of a video, the belongingness of the images to a panorama and to the neighbors is lost. However, a basic principle of the PatchMatch algorithm is the interaction of exploration through randomized hypothesis generation and propagation of good hypotheses to neighboring pixels. By processing individual cube faces independently, propagation is artificially interrupted at the boundary lines of the cube faces.

\subsection{View Filter}

The view filter has the task of deciding for a keyframe from the queue whether it should be added to the sliding buffer for keyframes to be densified. The decision is based on an evaluation of the presumed suitability of the keyframe for stereo reconstruction in conjunction with the currently most recent keyframe already in the sliding buffer. The View Filter component proved to be extremely useful, since the keyframes originating from the local mapping thread do have about enough stereo for localization and scene reconstruction from the point of view of SLAM parameter optimization in the context of a local map of neighboring keyframes. But just for a limited number of neighbors and especially at the beginning of the mapping, where the keyframe density is high and their distances are small.
Checking the stereo constraints of the keyframe candidate to the most recent keyframe in the buffer is done by SLAM using the sparse point cloud. Let:
\begin{equation}
\theta_{p} = \arccos ((t_{KA} - p)(t_{KB} - p), \sqrt{(t_{KA} - p)^{2}(t_{KB} - p)^{2}}), 
\end{equation}
$\forall{p \in AB}$, with $K_A$, $K_B$ are two keyframes and $t_{KA}$, $t_{KB}$ are the positions of $K_A$ and $K_B$ in world coordinates and $\theta_p$ $\in$ $[6^{\circ},60^{\circ}]$. If the relative number of 20\% of the key points exceeds the angle conditions than the keyframe is excepted. 

\subsection{Initialisation}

\begin{figure}
\centering
\includegraphics[width=0.49\textwidth]{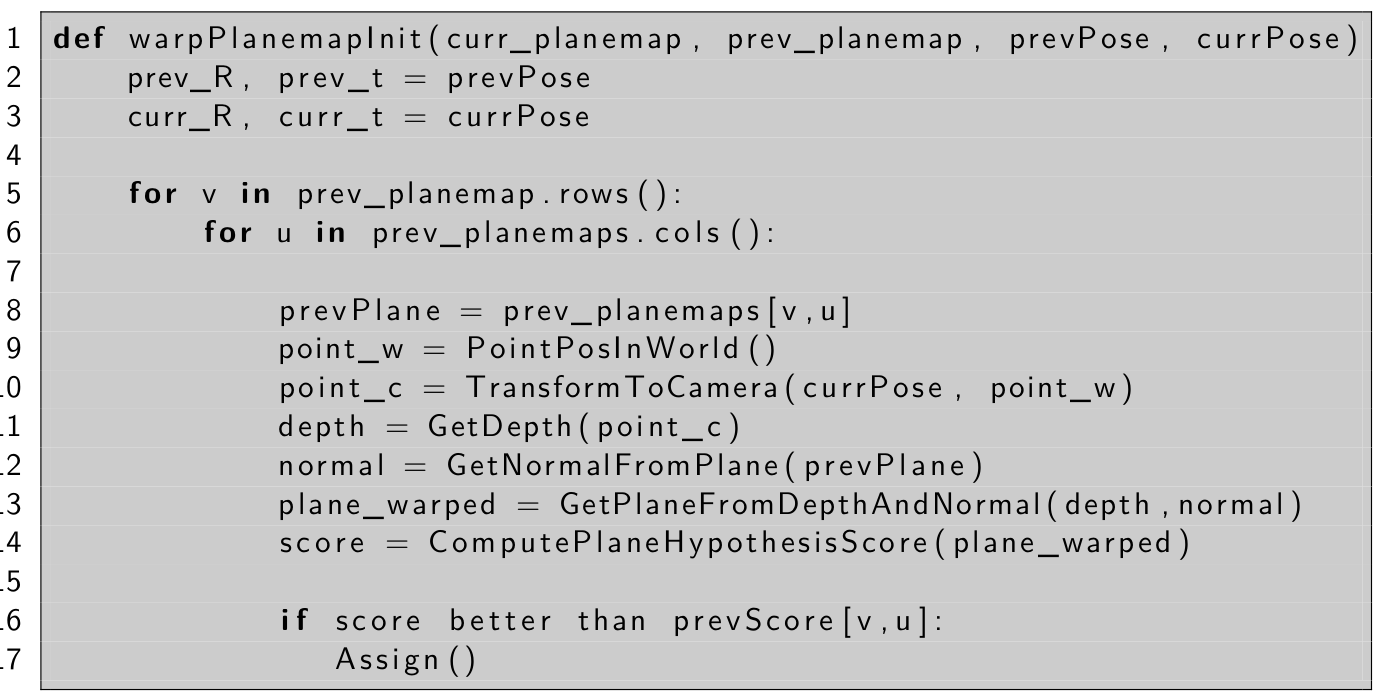}
\caption{Warp plane algorithm }
\label{algo:warp}
\end{figure}

Initialization of the plane map is a key component for the success of the process. The basis is the random initialization, which is common for PatchMatch-Stereo methods. However, due to the special circumstance of operating on a small keyframe sequence, this is supplemented by further initialization steps. The first one is plane warping. 
To reduce computation time, the stereo groups for densification are kept small. Thereby the images, even after view - filtering, often still contain a high degree of visual correspondence due to perspective correlation. Due to temporal proximity with respect to image acquisition, similar lighting conditions are to be expected, any motion blur due to abrupt movement of the camera or vibrations transmitted from the UAV to the camera are more likely to span adjacent keyframes of a temporally constrained sequence.
To counteract these drawbacks without increasing the number of frames in a stereo group,
the initialization of the plane map of the current keyframe can be done by transferring the optimization result of the preceding keyframe by plane map warping. This was already suggested by \cite{Shen2013AccurateMV} as a possibility for accelerated convergence (Figure. \ref{algo:warp}).

\subsection{Parallel 360° PatchMatch}

\begin{figure}
\centering
\includegraphics[width=0.49\textwidth]{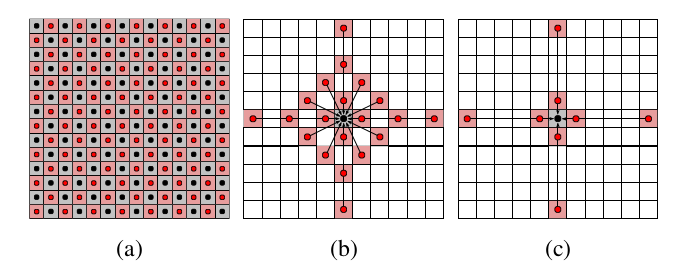}
\caption{Red-Black propagation scheme (a) Red and black pixels are each evaluated in parallel. For red pixels only black neighbors are evaluated and vice versa (b) A possible neighborhood for a black pixel. (c) Modified scheme for the accelerated execution \cite{7410463}.}
\label{fig:red-black-pattern}
\end{figure}

The implemented PatchMatch-Stereo algorithm differs from the classical only in the fact that a rectangular patch is selected in the reference keyframe, but it can assume any shape in the two neighboring frames due to the equirectangular projection. Furthermore, a parallel propagation scheme according to \cite{7410463} is used. Fig. \ref{fig:red-black-pattern} illustrates the scheme. For this work, an 8-neighborhood is used. Unlike in \cite{7410463}, however, the inner neighbors are shifted one position to the outside.
In addition, the following points are also be implemented:
\begin{enumerate}
    \item A random refinement routine is implemented, which is based on the PatchMatch-Stereo implementation of OpenSfM. A total of six random hypothesis tests are undertaken and with each step the search space is further restricted.
    \item In the course of post-processing, a median filter is applied to a copy of the depth image. The filtered image is compared with the original image. If the deviation is above a certain threshold value, the depth value is removed.
\end{enumerate}

\section{Evaluation}
\label{sec:eval}
\begin{figure}
\centering
\includegraphics[width=0.49\textwidth]{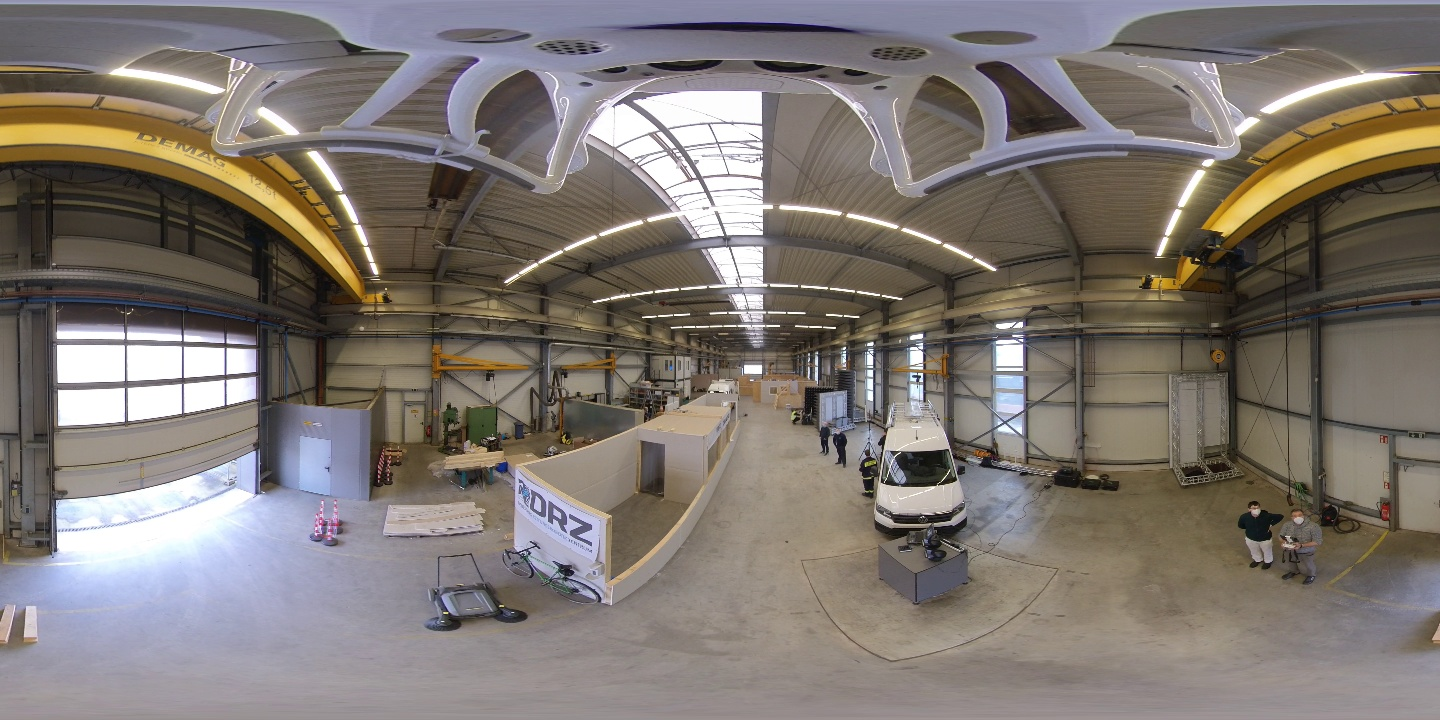}
\caption{360° view into the DRZ hall. }
\label{fig:drz-hall-360}
\end{figure}

\begin{figure}
\centering
\includegraphics[width=0.49\textwidth]{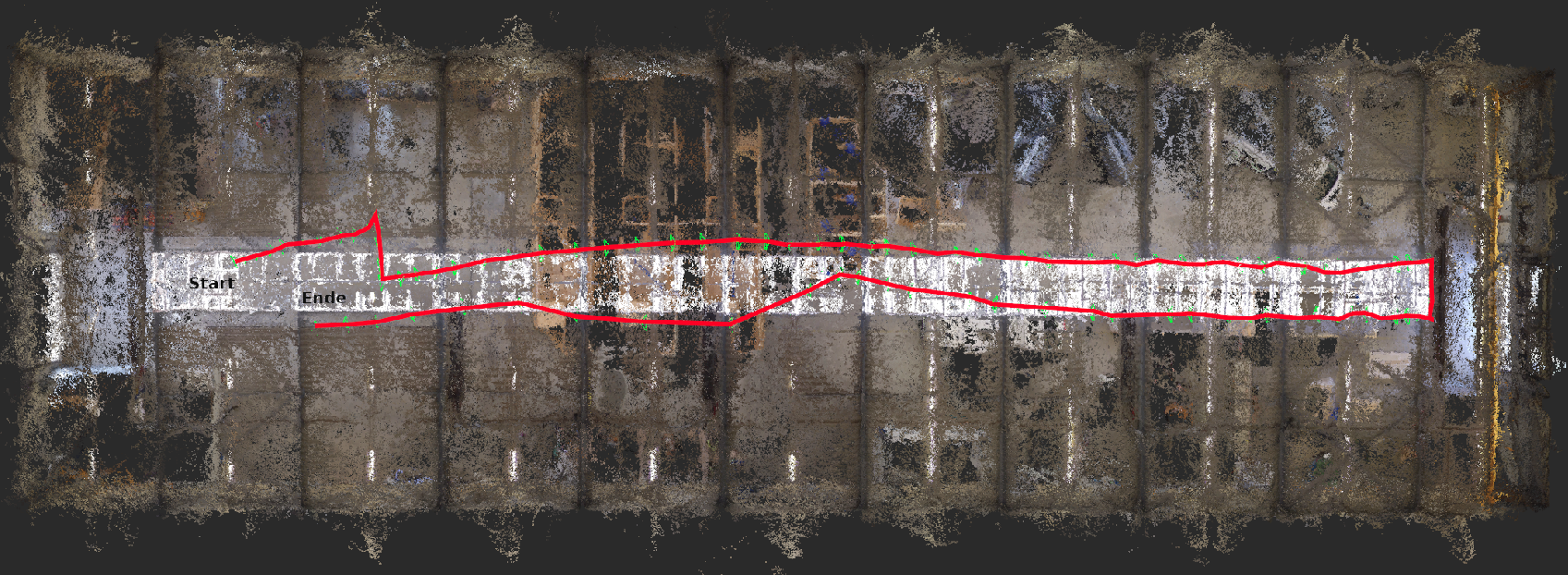}
\caption{Top view of the reconstructed point cloud and the trajectory of the UAV during a flight through the DRZ hall.}
\label{fig:drz-hall-traj}
\end{figure}

To evaluate the implementation we compare it on a sample flight in DRZ Hall (Fig. \ref{fig:drz-hall-360}). For this purpose we generate dense point clouds with OpenSfM\footnote{https://opensfm.org/} and OpenMVS\footnote{https://github.com/cdcseacave/openMVS}. Both methods split a 360° panorama into six frames and then apply the classical PatchMatch algorithm for pinhole cameras. In addition, we show the effect of plane warping, which does not play a role in the other methods. Besides the quality mode, we also show the result of an online run which is not possible with the other methods. The scenario corresponds to a rapid exploration of an industrial hall with a UAV equipped with a 360° camera. For this purpose, a UAV of the type DJI Phantom (Fig. \ref{fig:uav360}) is equipped with an Insta360 One X camera and flown in a straight line to the end of the hall at the German rescue robotics center in Dortmund - and just as straight line back again. This is intended to recreate the case of "quickly in and out again" (Fig. \ref{fig:drz-hall-traj}). The special feature of this scenario is that scene objects are primarily captured as they fly past; a more elaborate recording of a location in the scene from different angles is explicitly avoided, saving time and resources in data acquisition. Further examples in real rescue missions can be found here\footnote{A video of the dense mapping after a major fire of a an industrial hall with hazardous substance is given at (minute 3:00 ff): \url{https://www.youtube.com/watch?v=mR05-akD4BE}}.
The evaluation was done at a AMD Ryzen 9 5950X (16 cores, 64 GB RAM) with a NVIDIA Geforce RTX 2060 SUPER. The length of the video is 110 seconds.

\begin{figure}
\begin{center}
\includegraphics[width=0.23\textwidth]{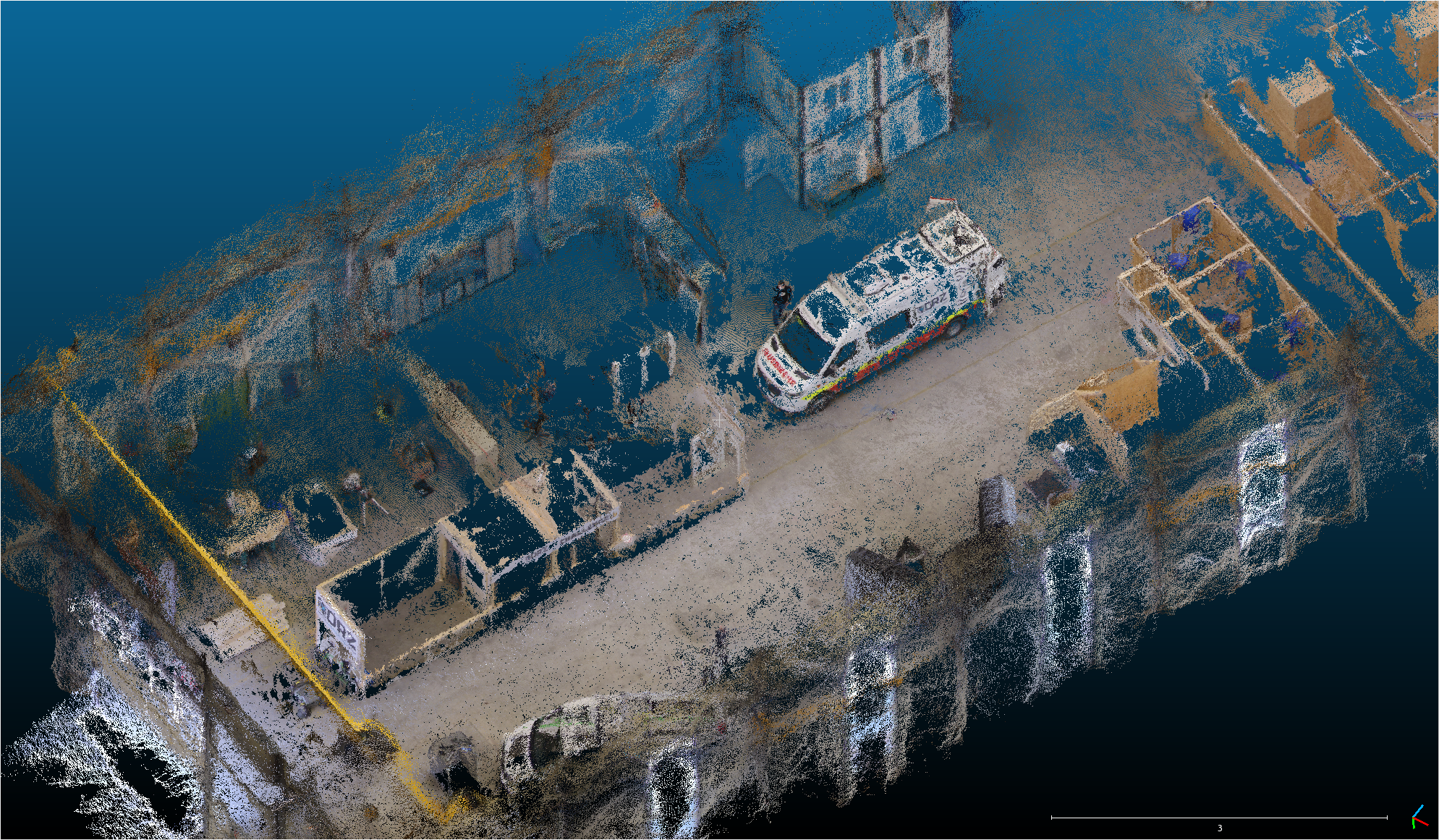}
\includegraphics[width=0.23\textwidth]{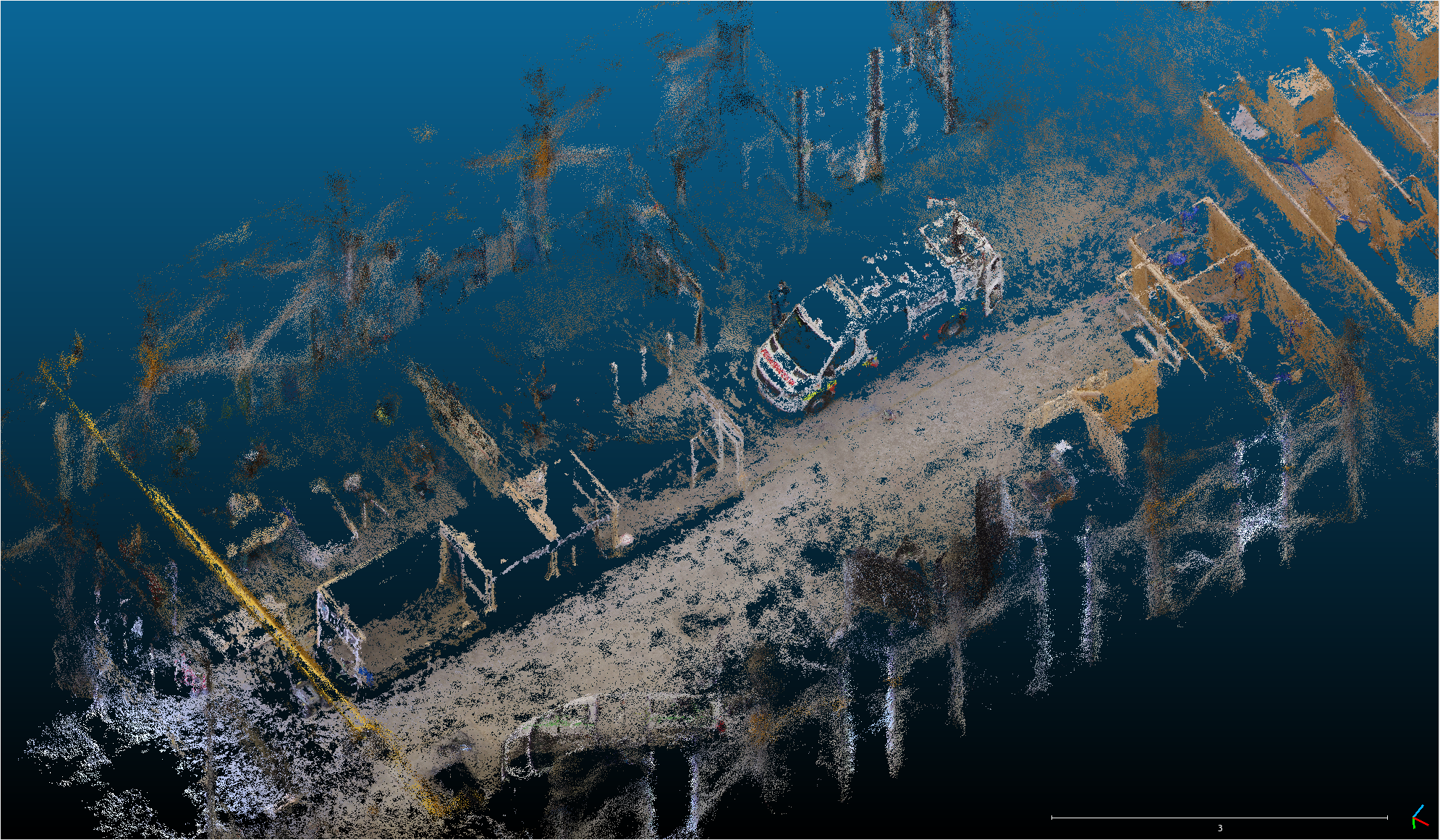}
\includegraphics[width=0.23\textwidth]{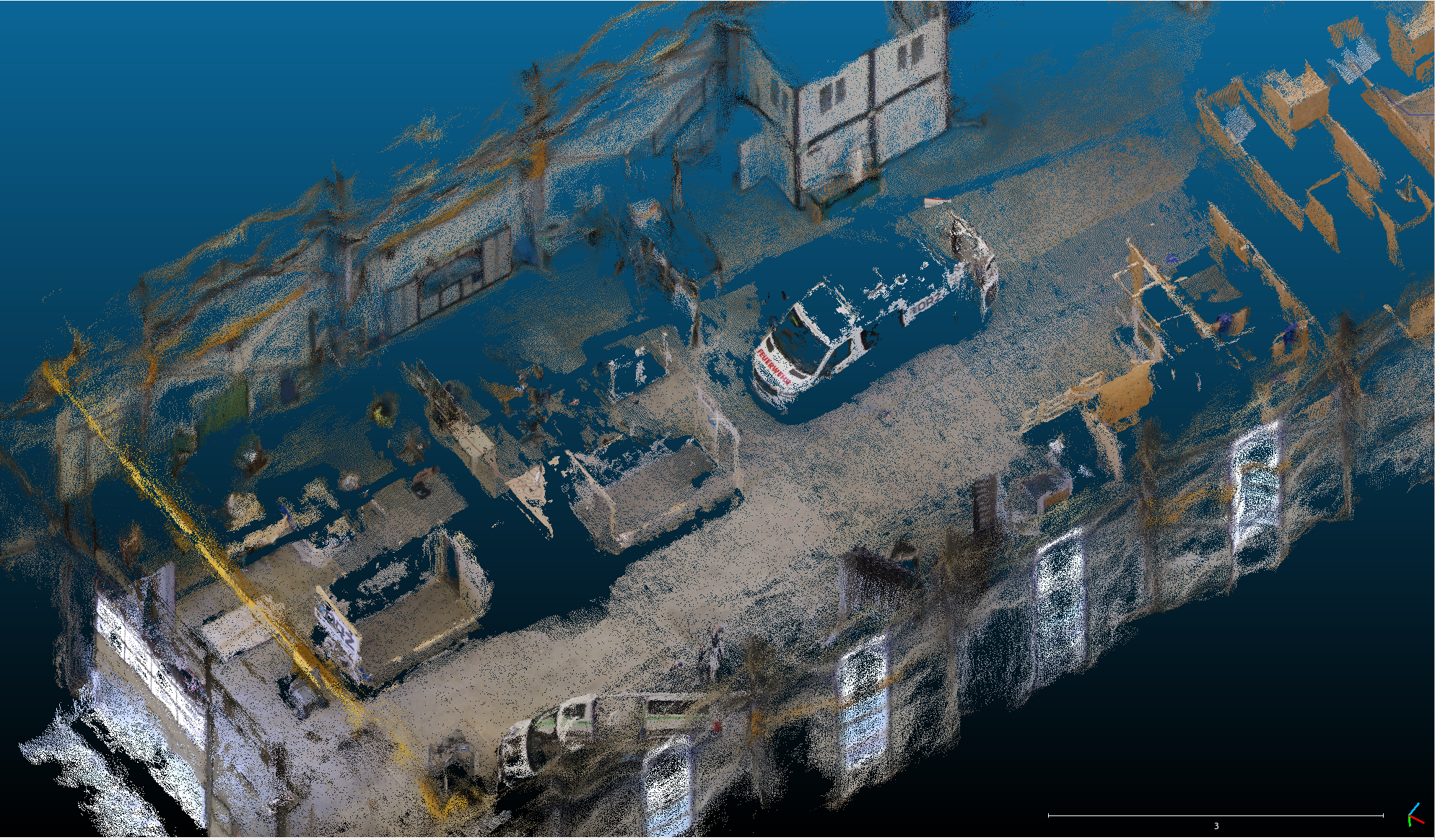}
\includegraphics[width=0.23\textwidth]{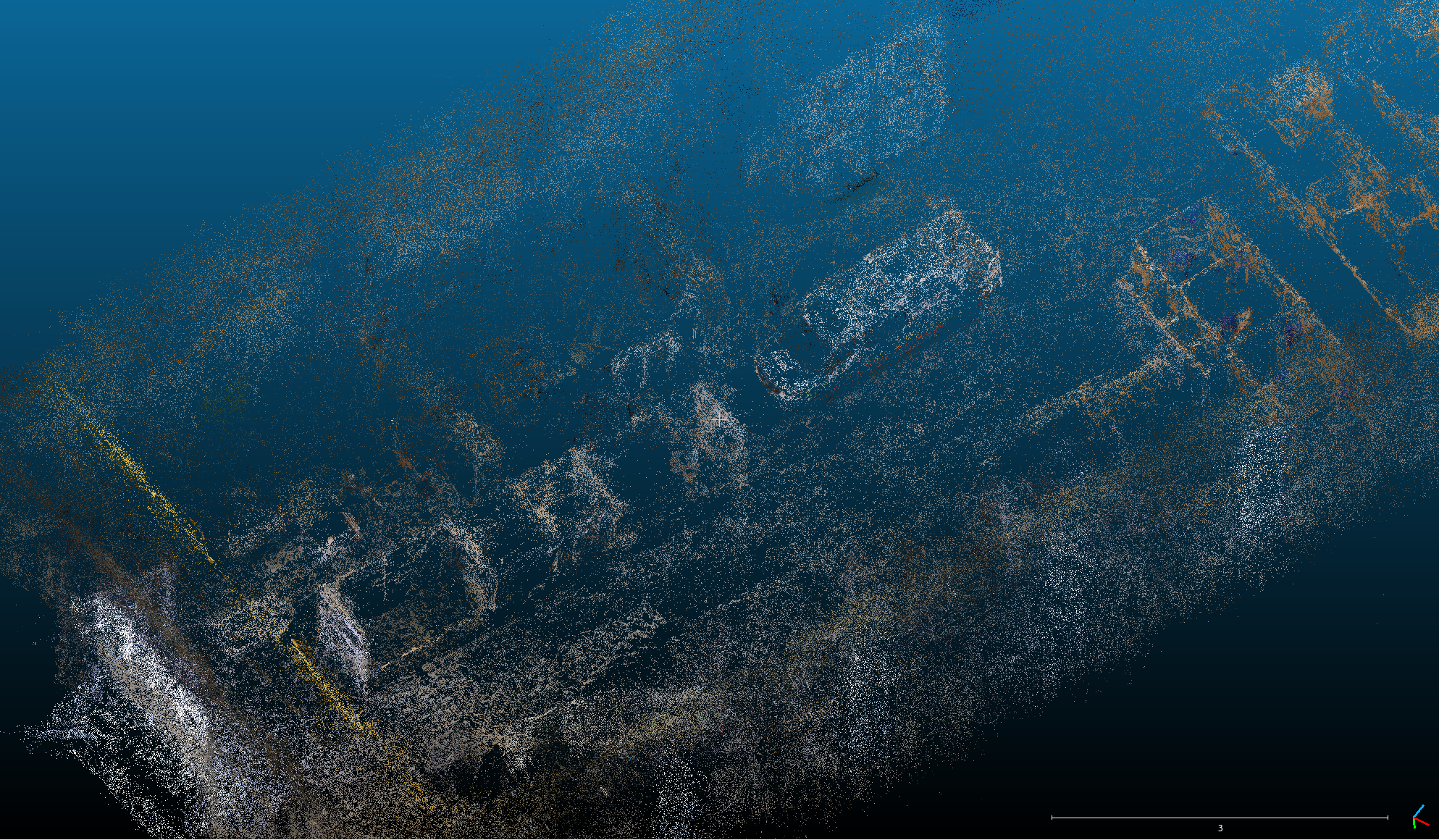}
\end{center}
\caption{3D Point clouds of the hall of the DRZ for the quality mode. Top left: PMD. Top right: PMD without initial warping. Bottom Left: OpenMVS. Bottom Right: OpenSfM }
\label{fig:drz-hall-res}
\end{figure}

\begin{figure}
\begin{center}
\includegraphics[width=0.23\textwidth]{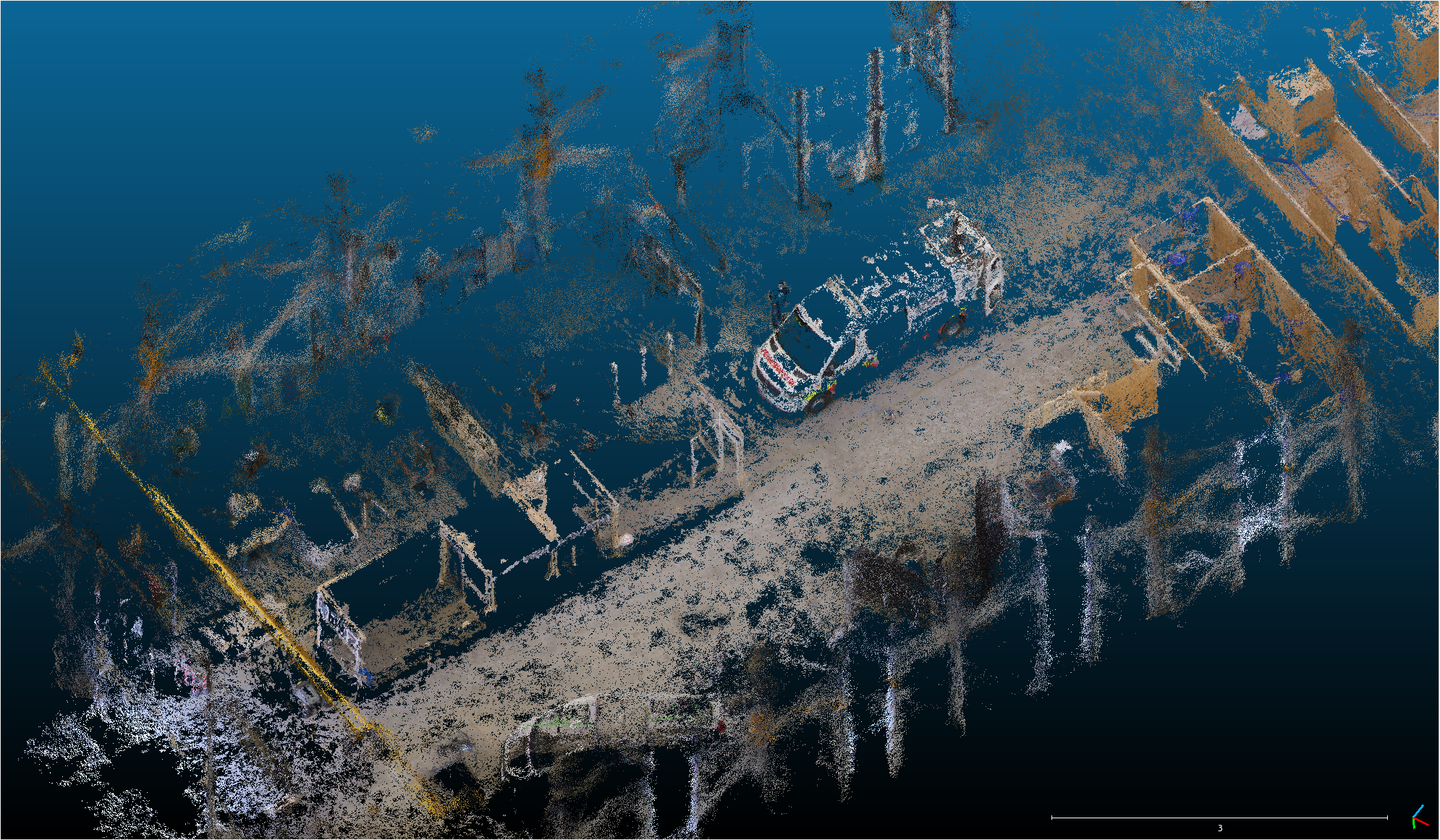}
\includegraphics[width=0.23\textwidth]{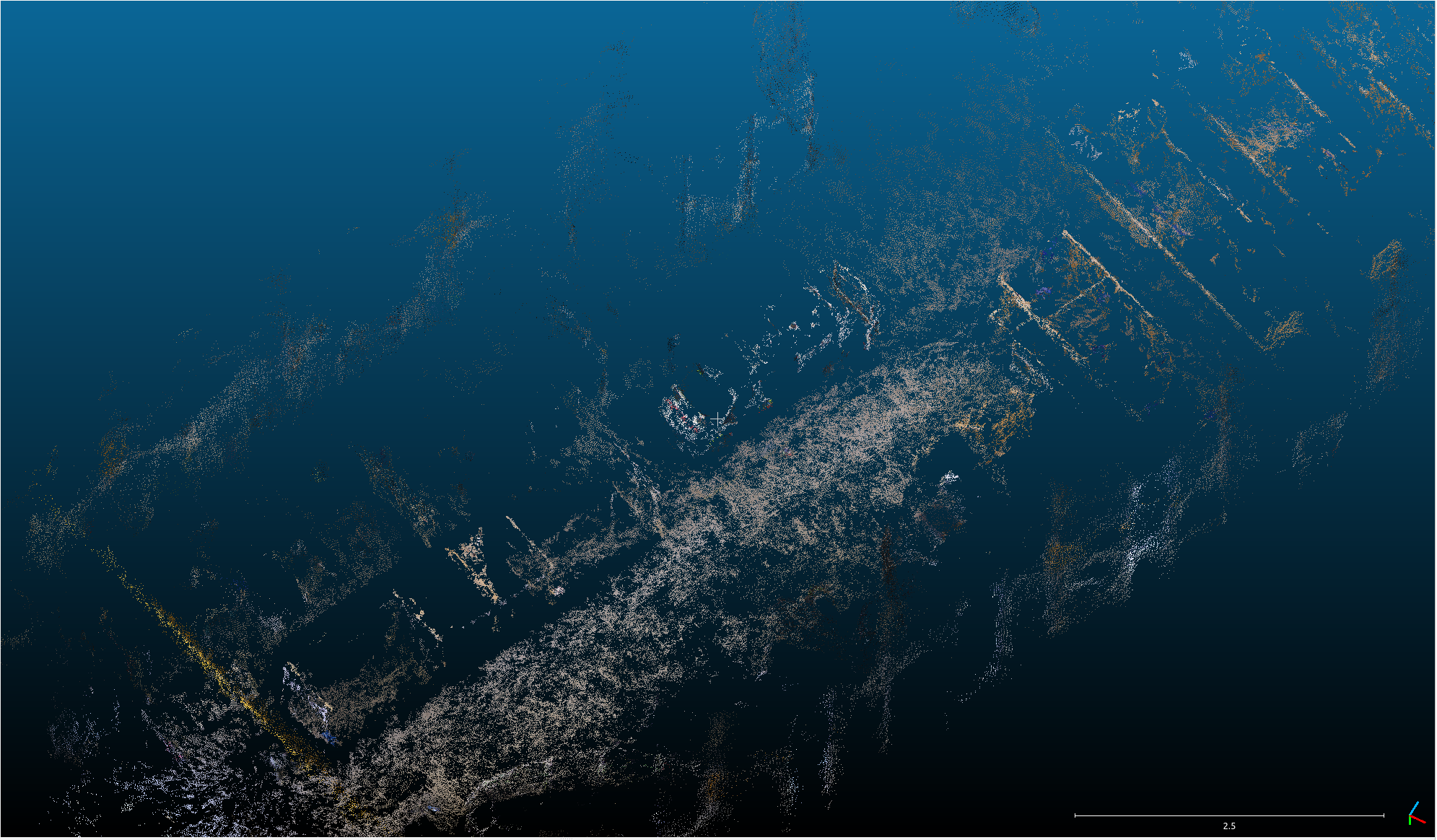}
\includegraphics[width=0.23\textwidth]{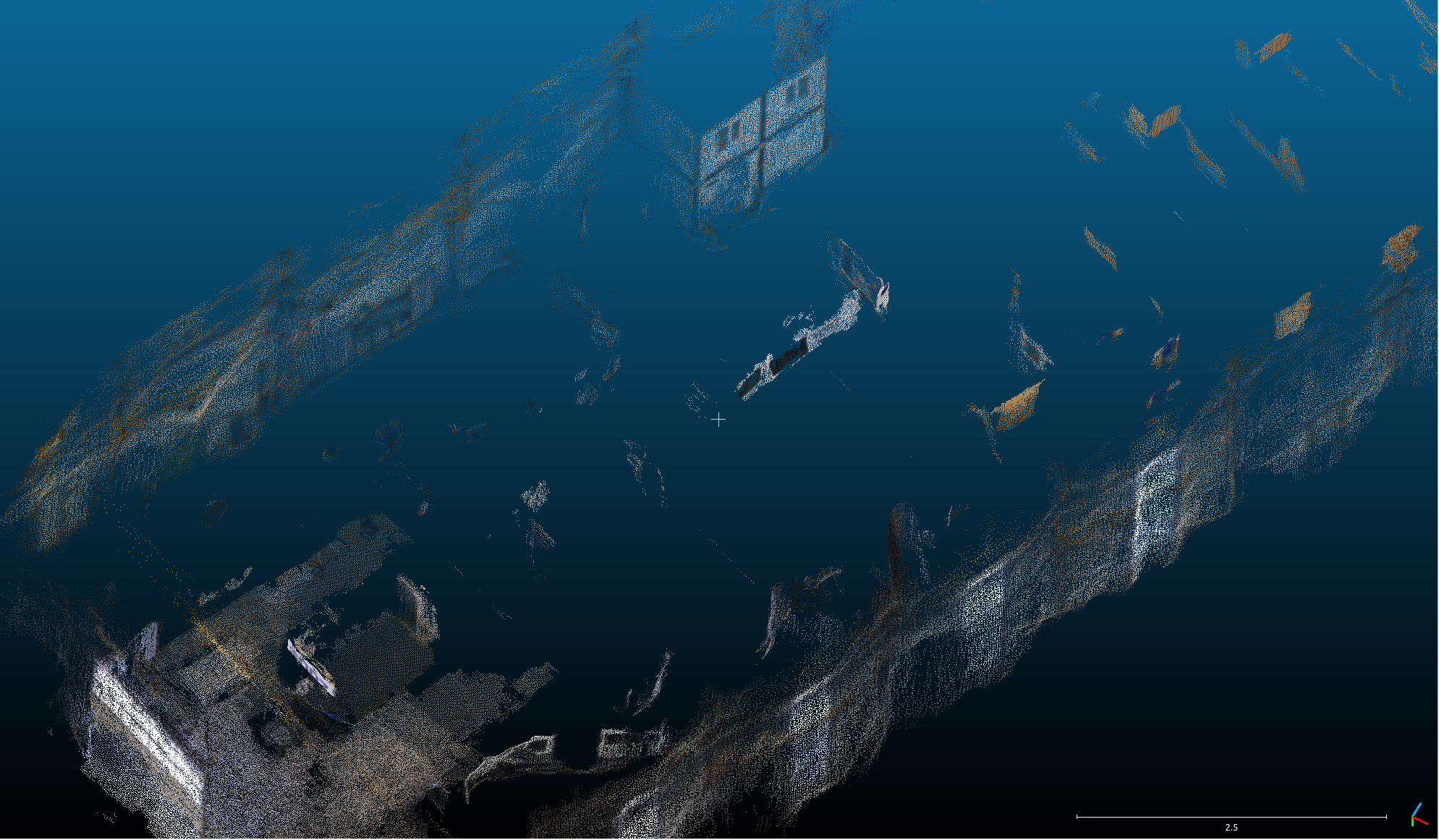}
\includegraphics[width=0.23\textwidth]{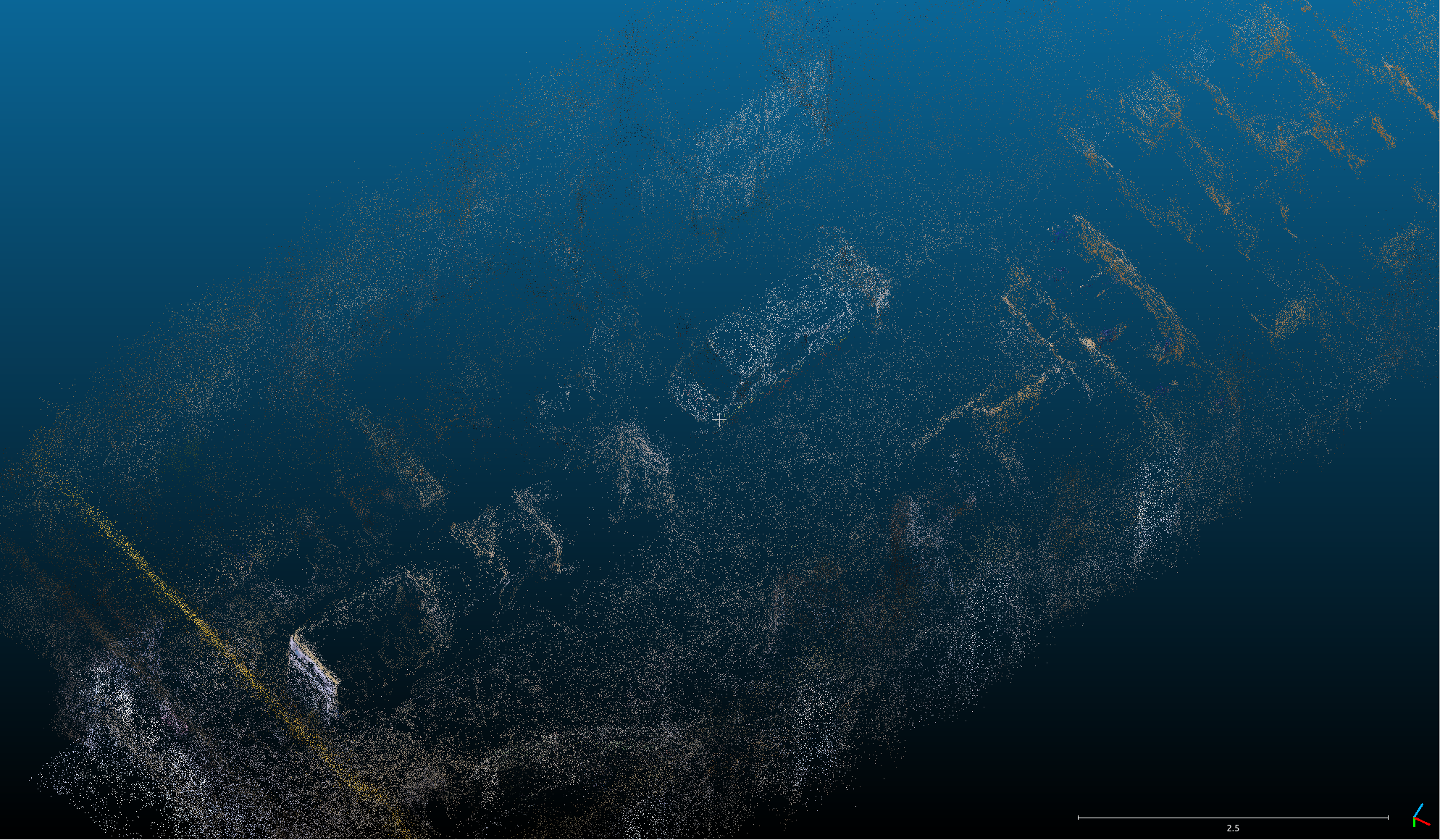}
\end{center}
\caption{3D Point clouds of the hall of the DRZ for the online mode. Top left: PMD. Top right: PMD without initial warping. Bottom Left: OpenMVS. Bottom Right: OpenSfM }
\label{fig:drz-hall-res-online}
\end{figure}

Figure \ref{fig:drz-hall-res} and Figure \ref{fig:drz-hall-res-online} compares the reconstruction by the developed method with warping to a reconstruction for which warping was deactivated. It can be clearly seen that the warping mechanism causes a significant increase in point density. Furthermore, a comparison with the point clouds by OpenMVS and OpenSfM shows that a higher completeness was achieved in the reconstruction of the scene (with active warping). In OpenMVS, the points are more homogeneous. OpenMVS has a special strategy for the post processing of the densification results, where point gaps are selectively filled. This could be the high density and homogeneous structure of the point cloud. The point cloud by OpenSfM is much sparser. For the online mode the video (30fps) has a resolution of 1920x960 and for the quality mode the full resolution of 5760x2880. In that mode the localization of OpenVSLAM goes down to $\sim$ 6Hz.

\begin{figure}
\begin{center}
\includegraphics[width=0.49\textwidth]{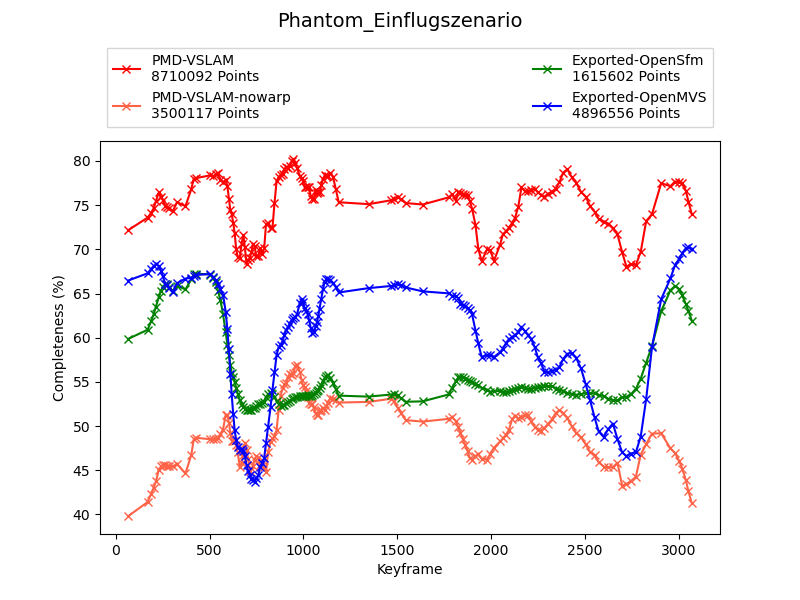}
\includegraphics[width=0.49\textwidth]{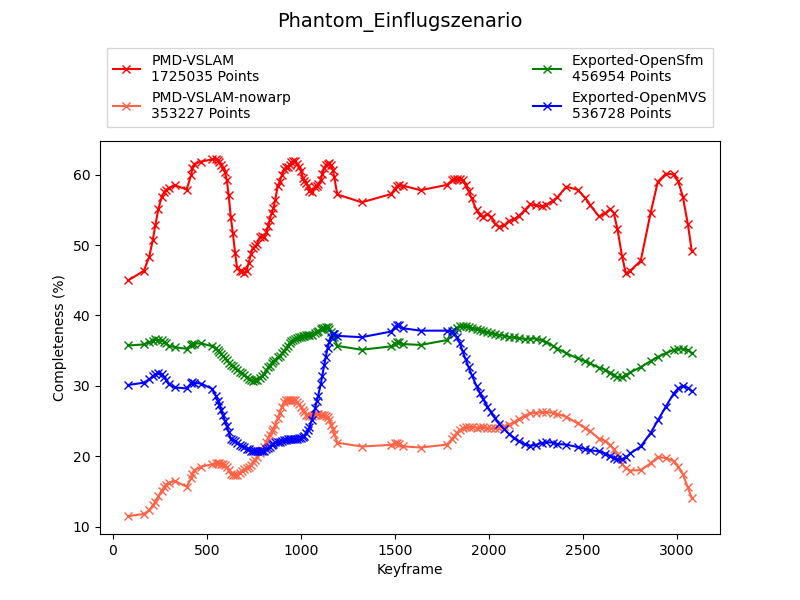}
\end{center}
\caption{Completeness of the point clouds of the 4 different approaches.}
\label{fig:drz-hall-abb-compl}
\end{figure}

Figure \ref{fig:drz-hall-abb-compl} confirms this impression quantitatively. A depth image is calculated at each keyframe position.  The ratio of the pixels on which a 3D point was projected and the total number of pixels contained.
This can be understood as a measure for local, keyframe related completeness of the point cloud. The depth panoramas are rendered at a resolution of 720x360 pixels.
For the PatchMatch-Stereo panorama with warping, the completeness is the highest of all comparison methods (\textasciitilde75\%) and the point cloud contains the most points (over 8 million). Without warping the method drops to the last place, the completeness is then only about 50\% and the
score drops to 3.5 million. Even fewer points were reconstructed by OpenSfM. The gain in completeness by warping is about 50\%, the gain in number of points is over 120\%. OpenMVS and OpenSfM position themselves in the middle in terms of completeness. OpenMVS is in front of OpenSfM in the middle video section, at the beginning and at the end both are about the same. OpenMVS has an advantage due to its ability to fill gaps in the point clouds. The number of points is higher by a factor of 3, but only reaches 50\% of the number of points of our online method.

Table \ref{tab:timing} shows the calculation time of the respective algorithms in online mode. For the localization we used the VSLAM time. OpenSfM and OpenMVS can also process unordered images, but then the computing time for the localization is an order of magnitude higher. Export and data conversion times are ignored. Since our densification is running together with OpenVSLAM no extra time is necessary.

\begin{table}[ht]
    \caption{Calculation time of the algorithms in online mode}
    \label{tab:timing}
    \centering
    \begin{tabular}{c|c|c|c}
      Mode: real time \\110s video & SLAM & densification & total \\ \hline \\
      PatchMatchDense (ours)  & 119s & +0 & 119s \\
      OpenSfM & 119s & +148s & 267s \\
      OpenMVS & 119s & +48s & 167s \\
    \end{tabular}
\end{table}

\section{CONCLUSIONS}
\begin{figure}
\begin{center}
\includegraphics[width=0.49\textwidth]{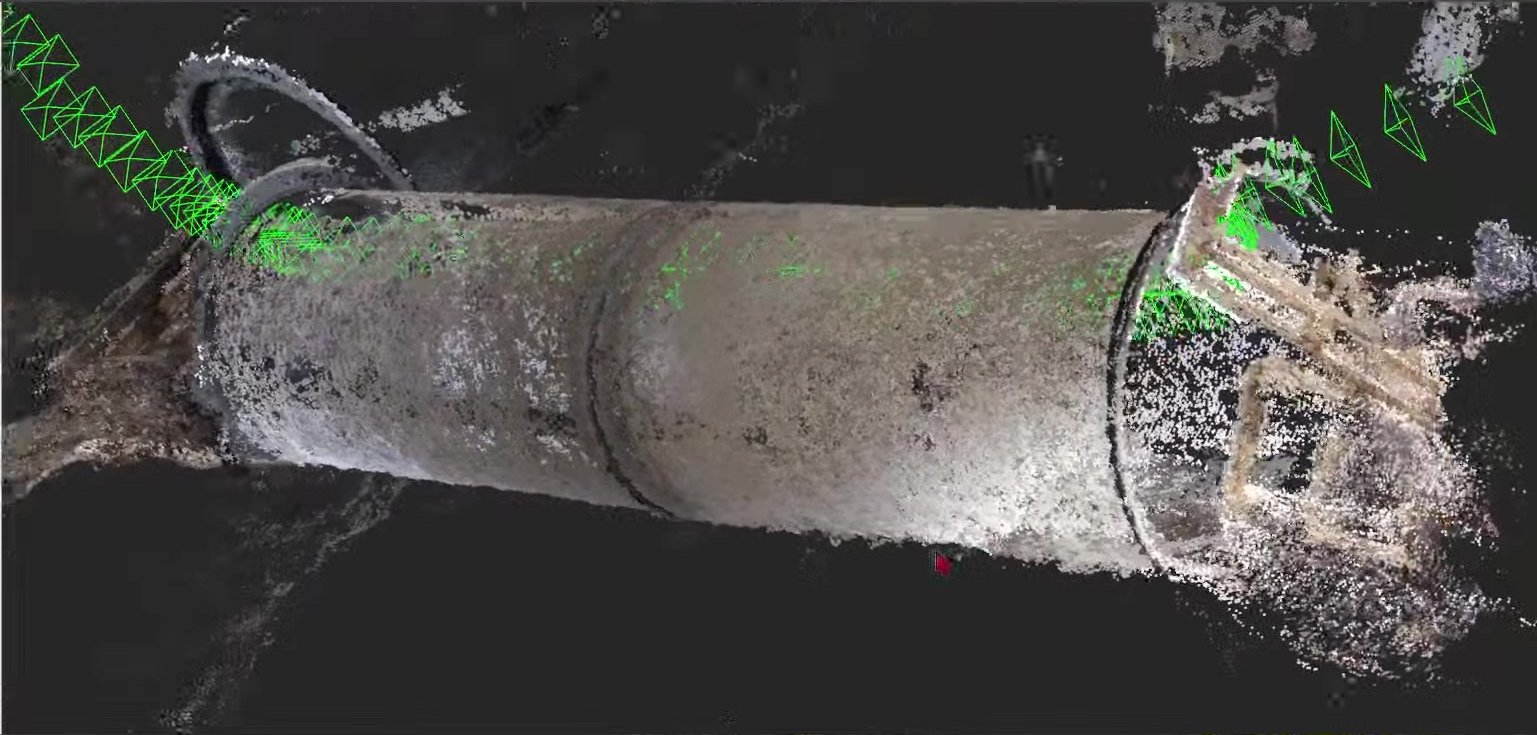}
\end{center}
\caption{Example of a flight through a pipe and the reconstructed 3D point cloud.}
\label{fig:drz-tube}
\end{figure}

We presented a real time dense SLAM system with a novel design that couples visual SLAM and PatchMatch Multi-View-Stereo for equirectangular images implemented on a graphic card. In particular, 
the use of 360° panoramic videos enables a robust and stable localization. Especially outside of research labs like in urban search and rescue environments, images with negative influence on the image quality like motion blur, over- or underexposure, noise or pronounced texture blur occur and prevent localization when only monocular cameras are used. The approach presented here manages the localization and the generation of a depth image even in a long uniform tube\footnote{\url{https://github.com/RoblabWh/PatchMatch}, \\ \url{https://www.youtube.com/watch?v=ybpNvSNzGto}} (Fig. \ref{fig:drz-tube}).

Needles to say a lot of work remains to be done. The PatchMatch-Stereo panorama determines dense correspondence relations already during SLAM localization. In addition to a possible application for online densification, these correspondences could serve as initial estimation for subsequent, more computationally intensive methods. Another interesting aspect with regard to the reuse of found correspondences is the possible reselection of selected images of the input video by relocalizing them based on the built map of the SLAM procedure and using them for the targeted improvement of a current densification result.

\section*{Acknowledgment}
This work was founded by the Federal Ministry of Education and Research (BMBF) under grant number 13N14860 (A-DRZ \url{https://rettungsrobotik.de/}. Thanks to all of our partners in the A-DRZ project).

\bibliographystyle{IEEEtran} 
\bibliography{IEEEabrv,literatur} 

\begin{thebibliography}{10}
\providecommand{\url}[1]{#1}
\csname url@rmstyle\endcsname
\providecommand{\newblock}{\relax}
\providecommand{\bibinfo}[2]{#2}
\providecommand\BIBentrySTDinterwordspacing{\spaceskip=0pt\relax}
\providecommand\BIBentryALTinterwordstretchfactor{4}
\providecommand\BIBentryALTinterwordspacing{\spaceskip=\fontdimen2\font plus
\BIBentryALTinterwordstretchfactor\fontdimen3\font minus
  \fontdimen4\font\relax}
\providecommand\BIBforeignlanguage[2]{{%
\expandafter\ifx\csname l@#1\endcsname\relax
\typeout{** WARNING: IEEEtran.bst: No hyphenation pattern has been}%
\typeout{** loaded for the language `#1'. Using the pattern for}%
\typeout{** the default language instead.}%
\else
\language=\csname l@#1\endcsname
\fi
#2}}

\bibitem{Kruijff2014}
\BIBentryALTinterwordspacing
G.~J.~M. Kruijff, M.~Jan{\'i}{\v{c}}ek, S.~Keshavdas, B.~Larochelle, H.~Zender,
  N.~J. J.~M. Smets, T.~Mioch, M.~A. Neerincx, J.~V. Diggelen, F.~Colas,
  M.~Liu, F.~Pomerleau, R.~Siegwart, V.~Hlav{\'a}{\v{c}}, T.~Svoboda,
  T.~Pet{\v{r}}{\'i}{\v{c}}ek, M.~Reinstein, K.~Zimmermann, F.~Pirri,
  M.~Gianni, P.~Papadakis, A.~Sinha, P.~Balmer, N.~Tomatis, R.~Worst,
  T.~Linder, H.~Surmann, V.~Tretyakov, S.~Corrao, S.~Pratzler-Wanczura, and
  M.~Sulk, \emph{Experience in System Design for Human-Robot Teaming in Urban
  Search and Rescue}.\hskip 1em plus 0.5em minus 0.4em\relax Berlin,
  Heidelberg: Springer Berlin Heidelberg, 2014, pp. 111--125. [Online].
  Available: \url{https://doi.org/10.1007/978-3-642-40686-7_8}
\BIBentrySTDinterwordspacing

\bibitem{Kruijff-amatrice}
I.~{Kruijff-Korbayová}, L.~{Freda}, M.~{Gianni}, V.~{Ntouskos}, V.~{Hlaváč},
  V.~{Kubelka}, E.~{Zimmermann}, H.~{Surmann}, K.~{Dulic}, W.~{Rottner}, and
  E.~{Gissi}, ``Deployment of ground and aerial robots in earthquake-struck
  amatrice in italy (brief report),'' in \emph{2016 IEEE International
  Symposium on Safety, Security, and Rescue Robotics (SSRR)}, Oct 2016, pp.
  278--279.

\bibitem{9738529}
H.~Surmann, D.~Slomma, R.~Grafe, and S.~Grobelny, ``Deployment of aerial robots
  during the flood disaster in erftstadt / blessem in july 2021,'' in
  \emph{2022 8th International Conference on Automation, Robotics and
  Applications (ICARA)}, 2022, pp. 97--102.

\bibitem{9597869}
I.~Kruijff-Korbayová, R.~Grafe, N.~Heidemann, A.~Berrang, C.~Hussung,
  C.~Willms, P.~Fettke, M.~Beul, J.~Quenzel, D.~Schleich, S.~Behnke,
  J.~Tiemann, J.~Güldenring, M.~Patchou, C.~Arendt, C.~Wietfeld, K.~Daun,
  M.~Schnaubelt, O.~von Stryk, A.~Lel, A.~Miller, C.~Röhrig, T.~Straßmann,
  T.~Barz, S.~Soltau, F.~Kremer, S.~Rilling, R.~Haseloff, S.~Grobelny,
  A.~Leinweber, G.~Senkowski, M.~Thurow, D.~Slomma, and H.~Surmann, ``German
  rescue robotics center (drz): A holistic approach for robotic systems
  assisting in emergency response,'' in \emph{2021 IEEE International Symposium
  on Safety, Security, and Rescue Robotics (SSRR)}, 2021, pp. 138--145.

\bibitem{9376551}
H.~Surmann, T.~Kaiser, A.~Leinweber, G.~Senkowski, D.~Slomma, and M.~Thurow,
  ``Small commercial uavs for indoor search and rescue missions,'' in
  \emph{2021 7th International Conference on Automation, Robotics and
  Applications (ICARA)}, 2021, pp. 106--113.

\bibitem{9597677}
H.~Surmann, D.~Slomma, S.~Grobelny, and R.~Grafe, ``Deployment of aerial robots
  after a major fire of an industrial hall with hazardous substances, a
  report,'' in \emph{2021 IEEE International Symposium on Safety, Security, and
  Rescue Robotics (SSRR)}, 2021, pp. 40--47.

\bibitem{Sumikura_2019}
\BIBentryALTinterwordspacing
S.~Sumikura, M.~Shibuya, and K.~Sakurada, ``Openvslam: A versatile visual slam
  framework,'' \emph{Proceedings of the 27th ACM International Conference on
  Multimedia}, Oct 2019. [Online]. Available:
  \url{https://arxiv.org/abs/1910.01122,
  https://dl.acm.org/doi/10.1145/3343031.3350539}
\BIBentrySTDinterwordspacing

\bibitem{6409456}
S.~Shen, ``Accurate multiple view 3d reconstruction using patch-based stereo
  for large-scale scenes,'' \emph{IEEE Transactions on Image Processing},
  vol.~22, no.~5, pp. 1901--1914, 2013.

\bibitem{Bleyer2011PatchMatchS}
M.~Bleyer, C.~Rhemann, and C.~Rother, ``Patchmatch stereo - stereo matching
  with slanted support windows,'' in \emph{BMVC}, 2011.

\bibitem{Mur_Artal_2015}
\BIBentryALTinterwordspacing
R.~Mur-Artal, J.~M.~M. Montiel, and J.~D. Tardos, ``{ORB}-{SLAM}: A versatile
  and accurate monocular {SLAM} system,'' \emph{{IEEE} Transactions on
  Robotics}, vol.~31, no.~5, pp. 1147--1163, oct 2015. [Online]. Available:
  \url{https://doi.org/10.1109%2Ftro.2015.2463671}
\BIBentrySTDinterwordspacing

\bibitem{Mur_Artal_2017}
\BIBentryALTinterwordspacing
R.~Mur-Artal and J.~D. Tardos, ``{ORB}-{SLAM}2: An open-source {SLAM} system
  for monocular, stereo, and {RGB}-d cameras,'' \emph{{IEEE} Transactions on
  Robotics}, vol.~33, no.~5, pp. 1255--1262, oct 2017. [Online]. Available:
  \url{https://doi.org/10.1109%2Ftro.2017.2705103}
\BIBentrySTDinterwordspacing

\bibitem{s21030705}
\BIBentryALTinterwordspacing
Y.~Zhang and F.~Huang, ``Panoramic visual slam technology for spherical
  images,'' \emph{Sensors}, vol.~21, no.~3, 2021. [Online]. Available:
  \url{https://www.mdpi.com/1424-8220/21/3/705}
\BIBentrySTDinterwordspacing

\bibitem{Campos_2021}
\BIBentryALTinterwordspacing
C.~Campos, R.~Elvira, J.~J.~G. Rodriguez, J.~M.~M. Montiel, and J.~D. Tardos,
  ``{ORB}-{SLAM}3: An accurate open-source library for visual,
  visual{\textendash}inertial, and multimap {SLAM},'' \emph{{IEEE} Transactions
  on Robotics}, vol.~37, no.~6, pp. 1874--1890, dec 2021. [Online]. Available:
  \url{https://doi.org/10.1109%2Ftro.2021.3075644}
\BIBentrySTDinterwordspacing

\bibitem{DSO}
J.~Engel, V.~Koltun, and D.~Cremers, ``Direct sparse odometry,'' \emph{IEEE
  Transactions on Pattern Analysis and Machine Intelligence}, vol.~40, no.~3,
  pp. 611--625, 2018.

\bibitem{7354184}
A.~Concha and J.~Civera, ``Dpptam: Dense piecewise planar tracking and mapping
  from a monocular sequence,'' in \emph{2015 IEEE/RSJ International Conference
  on Intelligent Robots and Systems (IROS)}, 2015, pp. 5686--5693.

\bibitem{10.1007/978-3-319-10605-2_54}
J.~Engel, T.~Sch{\"o}ps, and D.~Cremers, ``Lsd-slam: Large-scale direct
  monocular slam,'' in \emph{Computer Vision -- ECCV 2014}, D.~Fleet,
  T.~Pajdla, B.~Schiele, and T.~Tuytelaars, Eds.\hskip 1em plus 0.5em minus
  0.4em\relax Cham: Springer International Publishing, 2014, pp. 834--849.

\bibitem{6906584}
C.~Forster, M.~Pizzoli, and D.~Scaramuzza, ``Svo: Fast semi-direct monocular
  visual odometry,'' in \emph{2014 IEEE International Conference on Robotics
  and Automation (ICRA)}, 2014, pp. 15--22.

\bibitem{gao2018ldso}
X.~Gao, R.~Wang, N.~Demmel, and D.~Cremers, ``Ldso: Direct sparse odometry with
  loop closure,'' in \emph{iros}, October 2018.

\bibitem{odm2020}
O.~A. ODM, ``A command line toolkit to generate maps, point clouds, 3d models
  and dems from drone, balloon or kite images,''
  \url{https://github.com/OpenDroneMap/ODM}, 2020.

\bibitem{Shen2013AccurateMV}
S.~Shen, ``Accurate multiple view 3d reconstruction using patch-based stereo
  for large-scale scenes,'' \emph{IEEE Transactions on Image Processing},
  vol.~22, pp. 1901--1914, 2013.

\bibitem{7410463}
S.~Galliani, K.~Lasinger, and K.~Schindler, ``Massively parallel multiview
  stereopsis by surface normal diffusion,'' in \emph{2015 IEEE International
  Conference on Computer Vision (ICCV)}, 2015, pp. 873--881.

\end{thebibliography}

\end{document}